\pdfoutput=1
\documentclass{article}
\usepackage[font=small,labelfont=bf]{caption} 
\usepackage[utf8]{inputenc}

\usepackage{arxiv}

\title{Sequential Transfer Machine Learning in Networks: Measuring the Impact of Data and Neural Net Similarity on Transferability}

\author{
  Robin Hirt \\
  Karlsruhe Institute of Technology \\
  Karlsruhe, Germany \\
  \texttt{robin.hirt@kit.edu} \\
  \And
  Akash Srivastava \\
  MIT-IBM Watson AI Lab\\
  Cambridge, USA \\
  \texttt{akash.srivastava@ibm.com} \\
  \And
  Carlos Berg \\
  Karlsruhe Institute of Technology \\
  Karlsruhe, Germany \\
  \And
  Niklas Kühl \\
  Karlsruhe Institute of Technology \\
  Karlsruhe, Germany \\
  \texttt{niklas.kuehl@kit.edu} \\
}

\begin{document}
\maketitle

\begin{abstract}
In networks of independent entities that face similar predictive tasks, transfer machine learning enables to re-use and improve neural nets using distributed data sets without the exposure of raw data. As the number of data sets in business networks grows and not every neural net transfer is successful, indicators are needed for its impact on the target performance-its transferability.
We perform an empirical study on a unique real-world use case comprised of sales data from six different restaurants. We train and transfer neural nets across these restaurant sales data and measure their transferability. Moreover, we calculate potential indicators for transferability based on divergences of data, data projections and a novel metric for neural net similarity.
We obtain significant negative correlations between the transferability and the tested indicators. Our findings allow to choose the transfer path based on these indicators, which improves model performance whilst simultaneously requiring fewer model transfers.
\end{abstract}




\section{Introduction}

Machine learning is a main driver in the automation of process tasks across industries \citep{Sanders2016}. Although many industry players face similar problems with similar data structures in areas where machine learning can be utilized, every company typically solves these problems in an isolated manner \citep{Hirt2018}. From a systems perspective, these analytical tasks are well-comparable \citep{mizoguchi1995task}.

In an ideal world with an exhaustive exchange of all data across company borders, companies could solve similar problems in a more efficient manner \citep{Hirt2018}. However, due to competition and first and foremost, due to the preservation of intellectual property and privacy, a sharing of raw data is not feasible. From an economic standpoint, this poses a significant inefficiency as similar problems are solved multiple times and no analytical knowledge is exchanged. Additionally, the creation of analytical models is typically costly. If every company builds its own models, every company would end up with an inferior model as substantially more data potentially exists in the entire ecosystem. Moreover, every company would also have to reinvent the wheel, thus resulting in higher costs for model creation. Therefore, the current industry practices result in an inefficient resource utilization from a system's viewpoint \citep{hicks1939foundations}.

To address this challenge, we propose the utilization of transfer machine learning, a technique that enables to reuse and improve predictive machine learning models using different, distributed data sets. Hereby, no raw data exchange between companies is required, yet the transfer model can be improved by leveraging these different data sets. Although different types of analytical models could be transferred \citep{Hopf2017a}, neural networks are especially suited for transfer machine learning and are thus subject of the majority of related work \citep{Weiss2016}. Multiple studies demonstrate the effectiveness and efficiency of transfer machine learning in well-known, well-formed data sets like MNIST \citep{long2013transfer} or ImageNet \citep{huh2016makes}, but a lack of real-world industry studies is evident. One reason, amongst others, is the question on what, how, and when to transfer, since (naturally) not every neural net can be transferred to every data set \citep{Pan2009ALearning}. As our research gap, we observe a lack of techniques for identifying the impact of a neural net transfer prior to the transfer itself---which can be described as the transferability of a neural net. For the work at hand, transferability in general can be defined as the estimation of the extent to which representations learned from a source task can help in learning a target task \citep{Bao2019AnLearning}.  This is especially relevant when considering large numbers of participants in an ecosystem and a correspondingly high amount of potential neural nets candidates for transfer.

To address this gap, we perform an empirical study on a real-world use case with the aim to study the effects between different similarity measures and the transferability of neural nets. Precisely, we are interested in indicators for transferability of neural nets that are based on a comparison of data and data projections as well as on the neural nets themselves. As a basis for this study, we consider a unique data set of an ecosystem of different restaurant branches owned by different legal entities, all of whom need to perform sales forecasts to improve their respective resource allocations. As owners fear to expose data outside their restaurant, they are not willing to share raw data. Therefore, they are in need of a pre-transfer analysis on the possibility of value-adding neural nets without having to access the raw data of the competitor.

The paper at hand is structured as follows: In the remainder of this section, we cover related work, elaborate on our contribution to theory, define prerequisites and derive hypotheses. Then, we introduce the data set, present the neural net structure and the transfer, and elaborate on indicators for transferability based on raw data, data projections and neural nets. Afterwards, we present the results by first describing the performance impact of transferring neural nets in a business network. Then, we describe the impact of the tested indicators on transferability. After discussing our findings, we summarize the results, discuss their generalization, recognize limitations, and show future research prospects.

\subsection{Related Work and Contribution to Theory}

The foundations of transfer learning are surveyed by \citet{Pan2009ALearning} as well as \citet{Weiss2016} and provide a detailed overview on transfer learning. A wide variety of studies on the application of transfer learning can be identified: \citet{Zhong2010CrossLearning} present findings on the utilization of deep convolutional neural networks (CNN) in medical image analysis. They use large, general pre-trained sets and adapt them to a specific task to  show that pre-trained CNNs using computer vision databases (e.g., ImageNet) are useful in medical image applications and that multi-view classification is possible without the pre-registration of the input images. \citet{Kim2014ConvolutionalClassification} reports that pre-trained word vectors for sentence-level classification tasks can be seen as universal feature extractors that can be utilized for various classification tasks.
In this study, we focus on investigating the transferability \citep{Bao2019AnLearning} of neural networks from a source to a target domain. Related work can be divided into three main aspects that can indicate the transferability, namely the task similarity, the data similarity and, recently, also the model similarity. Table \ref{table:relatedTransferability} summarizes the related work on transferability in alignment with the aforementioned three main research categories.
A variety of work covering the topic of task similarity in transfer learning exists. \citet{Xue2007Multi-TaskPriors} classify tasks that are correlated and dependent, thus proving that concepts that were previously learned on one task may be transferred to other tasks. \citet{Yosinski2014HowNetworks} state that the transferability is negatively affected by the specialization of higher layer neurons of their source task, which eventually leads to a performance decrease on the target task. 
Another way to determine the transferability of neural nets is to examine the source and target data set itself. \citet{Jain2011} use the similarity among data points in order to update the detection score of the classifier and its classification boundary. \citet{Xiao2012a} find suitable training instances from other domains by measuring the distance between the source and target data in the domain of oil-prize forecasting. \citet{Zhong2010CrossLearning} apply density ratio weighting to overcome the difference in marginal distributions and propose a reverse validation procedure to quantify how well a neural net approximates the true conditional distribution of the target domain. However, there are more methods for comparing data distributions that could indicate transferability, such as divergences or distances \citep{Bhattacharyya1943OnDistributions, Eguchi1985AFunctionals, Kullback1951OnSufficiency}.

\begin{table}
\centering
\small
\begin{tabular}{|p{6cm}|p{2cm}|p{2cm}|p{2cm}|}
\hline
\textbf{Publication} & \textbf{Task \newline Similarity} & \textbf{Neural net\newline Similarity} & \textbf{Data \newline Similarity} \\ \hline
Multi-Task Learning for Classification with Dirichlet Process Priors \citep{Xue2007Multi-TaskPriors} & \multicolumn{1}{c|}{x} &  &  \\ \hline
How transferable are features in deep neural networks? \citep{Yosinski2014HowNetworks} & \multicolumn{1}{c|}{x} &  &  \\ \hline
SVCCA: Singular Vector Canonical Correlation Analysis for Deep Learning Dynamics and Interpretability \citep{Raghu2017SVCCA} &  & \multicolumn{1}{c|}{x} &  \\ \hline
Insights on representational similarity in neural networks with canonical correlation \citep{Morcos2018InsightsCorrelation} &  & \multicolumn{1}{c|}{x} &  \\ \hline
Cross Validation Framework to Choose Amongst Models and Datasets for Transfer Learning \citep{Jain2011} &  &  & \multicolumn{1}{c|}{x} \\ \hline
Online Domain Adaptation of a Pre-Trained Cascade of Classifiers \citep{Zhong2010CrossLearning} &  &  & \multicolumn{1}{c|}{x} \\
\hline
This work &  & \multicolumn{1}{c|}{x} & \multicolumn{1}{c|}{x} \\ \hline
\end{tabular}
\caption{Excerpt of related work on transferability and positioning of this work.}
\label{table:relatedTransferability}
\end{table}

Especially if the source data set is not available or cannot be accessed due to confidentiality reasons, examining a potential source neural network can be a way to gain insights on its transferability to a target data set. To the best of our knowledge, there is no work on finding indicators for transferability based on net structures. However, recent work shows possibilities for the comparison of neural net similarity using SVCCA \citep{Raghu2017SVCCA} to interpret neural network representations. \citet{Morcos2018InsightsCorrelation} apply SVCCA to compare net similarity across a group of CNNs, demonstrating that networks that generalize converge to more similar representations than networks that memorize.

In the course of this work, we are interested in transferring models across different data sets for which the data distribution may vary, but not the task to be executed. Thus, we disregard methods that are purely based on task similarity. We are interested in finding ways to receive indications on the transferability in a case where data cannot be pooled (e.g. due to confidentiality issues). To get an estimate of the basic indication of data similarity in transfer learning, we compare "raw" data sets. Then, in order to potentially reduce the amount of exposed information during the comparison, we examine ways to compare projections of those raw data sets. Given that even those projections might not be retrievable in some cases (e.g. in cases where only models are exchanged and initial training data is not accessible), we finally aim to find indicators for transferability based on the structure of a neural net.

The contribution of this work is threefold:
\begin{itemize}
\item We develop and evaluate the utility of a multi-step system-wide transfer on a unique data set in the domain of sales forecasting. 
\item We empirically show an association between the divergence of data distributions and the divergence of projection of data distributions with respect to the transferability of models. 
\item We empirically show that the Singular Value Canonical Correlation Analysis is associated with the transferability.
\end{itemize}

\subsection{Prerequisites and Research Design}

In our case, we want to transfer neural networks across different federated data distributions $p_l$ of $L$ companies:
\begin{equation}
\{p_l|l\in\{1,...,L\}\}.
\end{equation}
We define the input of $L$ different data sets $X^l$  that are composed of $B$ samples of a neural network $\eta$ as follows:
\begin{equation}
X^l=\{{x_i}\}^B_{i=0}|x^l\in \mathbb{R}^N.
\end{equation}
The test inputs $T^l$ and the corresponding true labels $V^l$ are composed of $h<B$ samples and are constructed by sampling uniformly from $X^l$:
\begin{equation}
T^l = \{{t_i}\}^h_{j=0}|V^l=\{{tr_j}\}^h_{j=0}|h<B.
\end{equation}
The performance M of a neural network $\eta_{p_l}$ trained on $p_l$ with predicted labels $\eta_{p_k} (X^l)$ is denoted as:
\begin{equation}
M(\eta_{p_l}(X^l),V^l)
\end{equation}
The performance delta $\Delta M$ of a source neural network $\eta_{p_k,p_z}$ which is trained on a distribution $p_k$ and then transferred to a target distribution $p_z$ is described as
\begin{equation}
\Delta M (\eta_{p_{k}},\eta_{p_{k},p_{z}}) = M(\eta_{p_{k},p_{z}}(X^z),V^z) - M(\eta_{p_z}(X^z),V^z)|\Delta M(\eta_{p_z},\eta_{p_k,p_z} \in \mathbb{R}^N.
\end{equation}
We define $\Delta M (\eta_{p_{k}},\eta_{p_{k},p_{z}})$ as the transferability of a model that is trained on the source distribution $p_k$ and transferred to the target distribution $p_z$. In our case, we therefore regard transferability as a performance increase of a neural network from one (source) distribution to another (target) distribution. The first goal of our work is to show that transferability, i.e. the performance increase of a transferred model, exists for the regarded problem/data set. Therefore, we formulate our first hypothesis as follows:

\textbf{Hypothesis 1 (H1):} A model $\eta_{p_k,p_z}$ which is pre-trained on a distribution $p_k$  and transferred to a distribution $p_z$ outperforms a model $\eta_{p_z} \iff \Delta (\eta_{p_z} ,\eta_{p_k,p_z})\geq 0$.

If this hypothesis can be confirmed, the next step of this work consists of identifying possible indicators for transferability in advance to the transfer itself. In order to do so, we analyze indicators for the transferability $\Delta M (\eta_{p_z},\eta_{p_k,p_z})$ by comparing $p_k$ and $p_z$ directly as well as their respective projections. Hereby, a projection $f$ maps $a$ distribution $p$ as follows: 
\begin{equation}
f:\mathbb{R}^a \longrightarrow  \mathbb{R}^b
\end{equation}
The projected distribution is $f(x_i)$. To empirically test different projections, we apply Multidimensional Scaling (MDS), Principal Components Analysis (PCA) and t-stochastic Neighborhood Estimation (t-SNE).
\begin{equation}
f\in {MDS,PCA,t-SNE}.
\end{equation}
To compare two distributions $p_k$ and $p_z$ and their respective projections we calculate their data divergence $D[p_k ||p_z]$ and data projection divergence  $D[f(x^k)||f(x^z)]$.

In this work, we aim to empirically examine the association between the divergence $D[p_k ||p_z]$ of data distributions $p_k$ and $p_z$, the divergence $D[f(x^k)||f(x^z)]$ of projected distributions $f(x^k)$ and $f(x^z)$ and the performance impact $\Delta M (\eta_{p_z},\eta_{p_k,p_z})$. Accordingly, we formulate Hypothesis 2 and 3:

\textbf{Hypothesis 2 (H2):} The divergence of two distributions $p_k$ and $p_z$, described as $D[p_k||p_z]$, correlates with the transferability $\Delta M(\eta_{p_z},\eta_{p_k,p_z})$.

\textbf{Hypothesis 3 (H3):} The divergence of the projection of two distributions $f(x^k)$ and $f(x^z)$, described as  $D[f(x^k)||f(x^z)]$ correlates with the transferability $\Delta M(\eta_{p_z},\eta_{p_k,p_z})$.

Finally, we examine neural nets themselves without accessing the source data to find indicators for transferability. Therefore, we consider the Singular Value Canonical Correlation Analysis (SVCCA). SVCCA enables the comparison of the behavior of neural nets, derived by the activations of neurons with regard to a data input $d^z$.  Let $\rho = (\eta_{p_k},\eta_{p_z },d^z)$ denote the result of an SVCCA between a net $\eta_{p_k}$ and a net $\eta_{p_z}$ based on a data sample $d^z \subseteq x^z$. Accordingly, we formulate Hypothesis 4:3

\textbf{Hypothesis 4 (H4):} The output of a Singular Value Canonical Correlation Analysis $\rho (\eta_{p_k},\eta_{p_z},d^z)$ correlates with the transferability $\Delta M (\eta_{p_z},\eta_{p_k},p_z)$.

\begin{figure}[H]
    \centering
    \includegraphics[width=14cm]{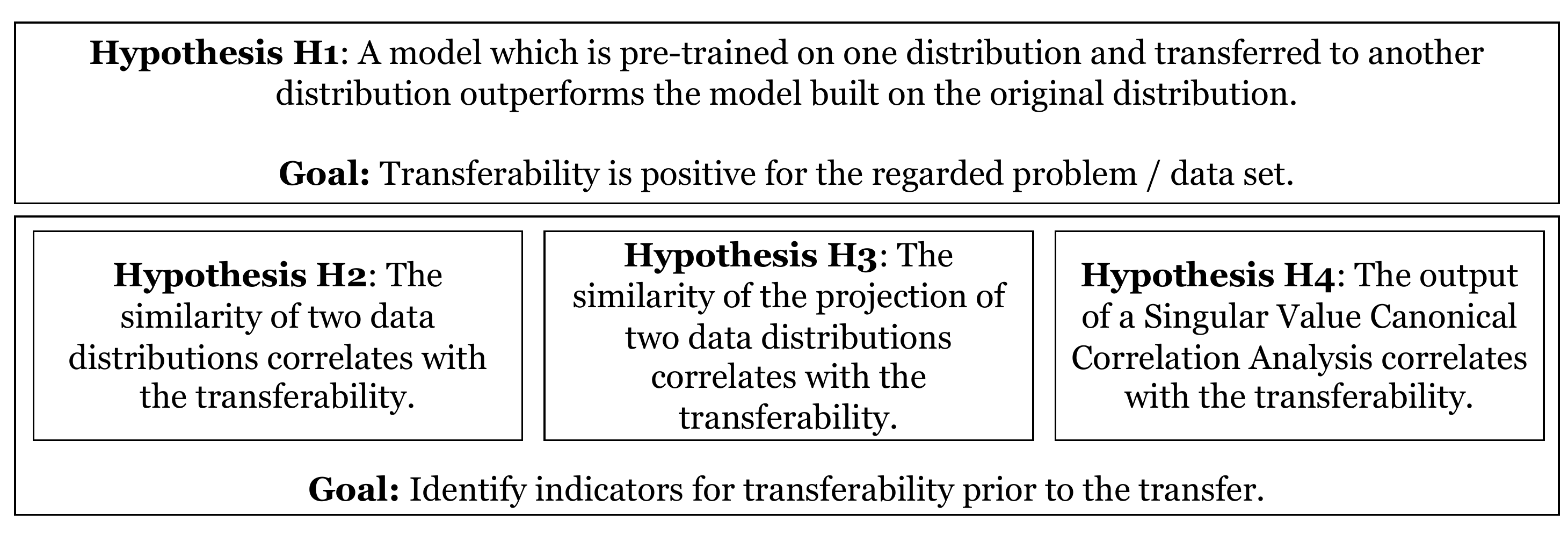}
    \caption{Overview of hypotheses and corresponding goal.}
    \label{fig:1-hypotheses}
\end{figure}

In Figure \ref{fig:1-hypotheses} we give an overview of our hypotheses and their corresponding goal. For H1, we perform a two-sided one-sample t-test for the mean of all transferabilities to test if the average transferability significantly deviates from zero. For H2-4, we calculate Spearman's rank correlation coefficient as a non-parametric measure between the variables and test the significance of the calculated Spearman's rho $r_s$.

\section{Experiment}

In this chapter, we first give an overview of the data we examined and subsequently elaborate on the sales forecasting model design and the transfer mechanism. In conclusion, we describe how we compare data and data projections. Lastly, we present the applied variation of measuring the net similarity via SVCCA.

\subsection{Data Set}

We analyze unique daily sales data of six different restaurant branches of two particular restaurant chains that serve different types of food. The data set captures observations from 2013 until 2017. 

\begin{table}[htp]
\centering

 \begin{tabular}{rrrrrrrr}
  \hline
Branch & 1 & 2 & 3 & 4 & 5 & 6 \\ 
  \hline
  Company & A & A & A & B & B & B \\ 
  City & a & a & b & a & c & d\\ 
  \hline
\end{tabular}
\caption{Overview of available data for branch 1 to 6 (sales data from 2013-01-01 to 2017-12-31).}

\label{tab:data-overview-tl}
\end{table}

By precisely predicting the sales per day for each branch in the next week, month, or even year, several advantages can be leveraged: based on the revenue and demand, staff schedules can be optimized toward cost savings and a better experience for customers can be delivered. Additionally, the procurement of supplies can be improved, as spoiled food is a main cost-driver for restaurants. Thus, the management of restaurant chains has a major interest to forecast sales for their branches.

Table \ref{tab:data-overview-tl} gives an overview on all the available branch data we use in this work. Each of the two restaurant companies has three branches with different locations. Branch 1, 2 and 4 are located within the same city.

\begin{figure}
    \centering
    \includegraphics[width=12cm]{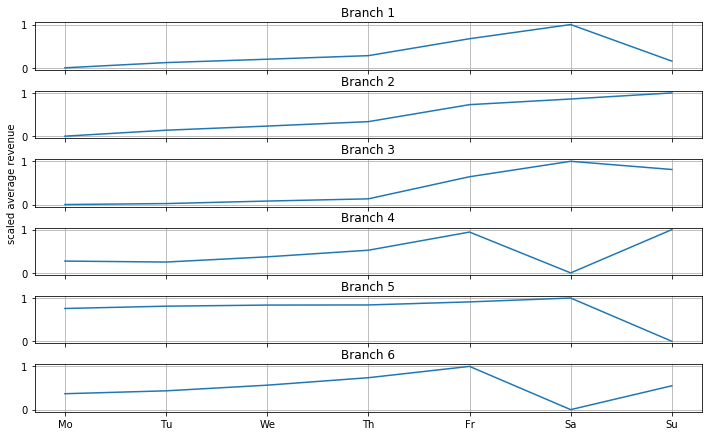}
    \caption{Average branch revenue over days of the week, scaled.}
    \label{fig:2-tl}
\end{figure}

Figure \ref{fig:2-tl} compares the average weekly revenue of each branch. We can recognize a different weekly seasonality for the revenue of the restaurants. Branches 1, 2 and 5 have their highest sales on Saturday, while branch 4 and 6 reach their minimum on that very day. Different market orientation, opening hours and locations are possible reasons for this observation. Hence, branch 5 appears to be closed on Sundays. In general, a restaurant in the commercial city center can attract more customers on the weekend than one that is located in an industrial area of town. In those areas, offices or production businesses are located which tend to be closed on those days. All branches, with the exception of the aforementioned two branches, share the common behavior with an uptrend in net revenue starting from Monday and reaching their peak over the weekend.

\subsection{Sales Forecasting Model Design and Transfer}

We aim to build separate models for each data distribution, where one data distribution corresponds to the data set of one branch. Afterwards, these models are transferred to every other distribution and then re-trained. This procedure is repeated until every model has passed through every distribution exactly once (H1). To empirically study the effects of data, data projection divergence and net similarity on the transferability of models, we test all possible transfers in a brute-force attempt and analyze the results a posteriori (H2-H4).

Our goal is to develop a model that is able to forecast daily sales on a weekly basis. There are many ways to design a sales forecasting model, such as ARIMA models, additive, or logarithmic regressions. To simplify our research design, we focus solely on Convolutional Neural Networks (CNN) for multivariate forecast as they have proven to achieve superior results in similar problems in the past \citep{Borovykh2017ConditionalNetworks}. Here, the input $X_i$ of a neural network $\eta$ is:

\begin{equation}
X_i=  \{s_i^n,y_i,m_i,w_i \},     
\end{equation}

where $s_i^n$ is a vector of daily sales of the previous sales period, $y_i \in \mathbb{Z}$  denotes the year, $m_i \in Z$ the month and $w_i \in \mathbb{Z}$  the week of the observation. The complete data set can be described as $\{X_i\}_{i=0}^{B \times L}$   and $s_j^1 \in \mathbb{R}^7$. Then, the date and time index are adjusted and reformatted in line with the opening hours of the respective branches. The available variables are grouped by day in order to forecast the time series on a per day basis. We clean obvious errors in the data set by dropping erroneous values, such as negative daily revenues. 

\begin{figure}[H]
    \centering
    \includegraphics[width=12cm]{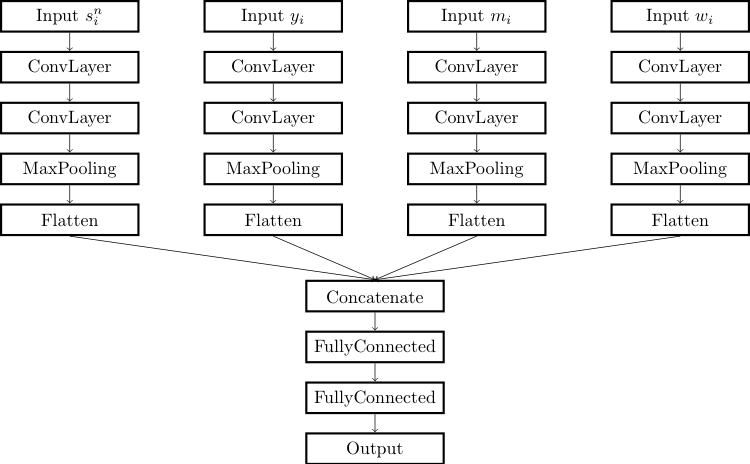}
    \caption{Multi-head architecture of employed CNN model.}
    \label{fig:3-multihead}
\end{figure}

As a next step, we build a multi-head CNN model to forecast the daily sales of the next sales period. The structure of the CNN model is depicted in Figure \ref{fig:3-multihead}. The model has four input variables: revenue of the previous sales period, month, weekday and year of the observation. Each variable is fed into a separate head. All heads consist of two one-dimensional convolutional layers with the same parameter configuration, followed by a max-pooling layer. The output of the pooling layers is flattened and merged by a concatenation layer. The merged heads’ output is fed into a first fully connected layer followed by a second one to conduct the interpretation. Finally, the sales forecast for the next period is generated in form of an output vector.

In a pre-test, we determine the model hyperparameters by empirical testing as follows: the two one-dimensional convolutional layers both have 32 filter maps and a kernel size of 3. As activation function, rectified linear unit is applied to both convolutional layers. The pool size for the max-pooling layer is set to 2. The first fully-connected layer contains 200 neurons and the second one 100 neurons. The model is compiled with mean squared error (MSE) as loss function during training and uses Adam as optimizer \citep{Kingma2015Adam:Optimization}. After compilation the model is fitted on the training data set for 20 epochs with a batch size of 16.

For the model training and re-training, we split the data into a training and a test set for each branch. As testing period, we choose the year 2017 consistently. The remaining data builds our training or re-training set. For every model $\eta_{p_x}$, we calculate its performance on the actual target data set and on the union of all test sets across all branches for comparability reasons. 

\begin{figure}
    \centering
    \includegraphics[width=7cm]{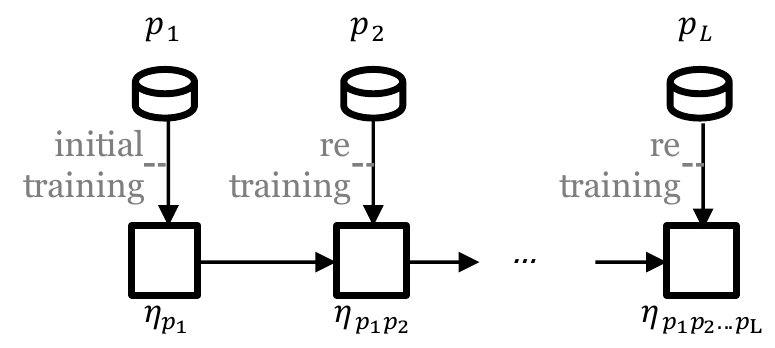}
    \caption{Overview of a possible transfer path for a model across different data distributions.}
    \label{fig:4-transferpath}
\end{figure}

To implement the transfer, we re-train a source CNN on a target data set as depicted in Figure \ref{fig:4-transferpath}. Hereby, we do not freeze the layers to enable re-weighting of the neurons in the layers. We re-train the CNN model with the same number of epochs (25) and batch size (16) as in base model training. Note that it would also be possible to adaptively choose certain layers to freeze and dynamically adapt the learning rate. For this study, we chose not to change or vary the amount of training parameters or frozen layers for a transfer. By choosing not to do so, the models are more likely to "forget" previously learned knowledge. Future work needs to address a more adaptive learning strategy. The degree of transfer denotes the total amount of performed transfers per model. In Table \ref{tab:possible_transfers} we give an overview of all transfers, their respective source models and the respective targets according the degree of transfer. Generally, the amount of transfers grows significantly with a growing number of data sets N and can be described by $\sum_{k=0}^{n-1}$ $\frac{n!}{k!}$.

\begin{table}
\centering
\begin{tabular}{rrrrrrrr}
  \hline
Degree of transfer & 1$^{st}$ & 2$^{nd}$ & 3$^{rd}$ & 4$^{th}$ & 5$^{th}$ & Total \\ 
  \hline
  Source models & 6 & 30 & 120 & 360 & 720 & 120 \\ 
  Possible targets & 5 & 4 & 3 & 2 & 1 & -\\ 
  Targets & 30 & 120 & 360 & 720 & 720 & 1950\\ 
  \hline
\end{tabular}
\caption{Number of possible transfers.} 
\label{tab:possible_transfers}
\end{table}

\subsection{Data and Data Projection Divergence}

In the following, we first introduce the utilized data divergence measure, which we apply on the unchanged data populations as well as on the projected data.
Measuring the independence or divergence of two random variables or distributions can be conducted in different ways. In this work, we estimate the divergence of two data distributions using an energy distance meta estimator $D_{EnDist}(f_1,f_2)$  as equivalent to maximum mean discrepancy \citep{Szabo2014InformationToolbox, Szekely2013EnergyDistances}, which is defined as follows:
\begin{equation}
D_{EnDist} (f_1,f_2 )= 2[D_{MMD} (f_1,f_2)]^2
\end{equation}
\begin{equation}
D_{MMD} = ||\int \{p_kK(;f(x^k)) \delta f(x^k) - \int p_zK(;f(x^z)) \delta f(x^z) \}||^2_\mathbb{H}
\end{equation}

Considering a scenario where data cannot be exchanged across entities of a system, it is not possible to compare two data sets simultaneously. To ensure a certain degree of confidentiality, a possible solution would be to compare only projected data, where critical information is already lost due to abstraction \citep{Narayanan2008}. 

\begin{figure}
    \centering
    \includegraphics[width=12cm]{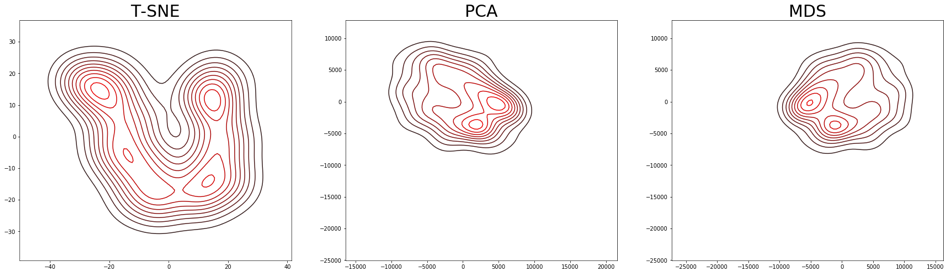}
    \caption{Bi-variate kernel density estimates of data projection (t-SNE, PCA, MDS) for data distribution $\delta_1$ of the first branch.}
    \label{fig:5-kde}
\end{figure}

Thus, in an initial step we apply projections $f:\mathbb{R}^a \rightarrow \mathbb{R}^b$  raw data $\delta_x \in \mathbb{R}^a$ to retrieve abstractions $\delta_x \in \mathbb{R}^b$ where $a>b$. We use three established algorithms to calculate abstractions of the raw data, namely t-distributed stochastic neighbor embedding (t-SNE), multidimensional scaling (MDS) as well as principal component analysis (PCA). The t-SNE is a well-suited technique for the visualization of high-dimensional data to create meaningful intermediate results and is effective for interactive data analysis \citep{Pezzotti2017ApproximatedAnalytics}. MDS is a technique used for analyzing similarity or dissimilarity of data. It attempts to model the relationship between data as distances in a geometric space \citep{Borg2003ModernApplications}. Lastly, PCA decomposes a multivariate data set into a set of subsequent orthogonal components which explain a maximum amount of the variance in the data \citep{Halko2011FindingDecompositions}. The projections for each technique applied to the first data distribution $\delta_1$ are visualized in Figure \ref{fig:5-kde}.

Subsequently, we calculate divergences between the data projections. Lastly, for both the raw data and data projections, we evaluate whether a correlation to the transferability of models is given.

\subsection{Neural Net Similarity}

The Singular Value Canonical Correlation Analysis (SVCCA) is a method for analyzing and comparing different representations learned by artificial neural networks \citep{Raghu2017SVCCA}. It represents an amalgamation of a singular value decomposition (SVD) and a canonical correlation analysis (CCA) \citep{Hardoon2004CanonicalMethods}. 

\begin{figure}
    \centering
    \includegraphics[width=14cm]{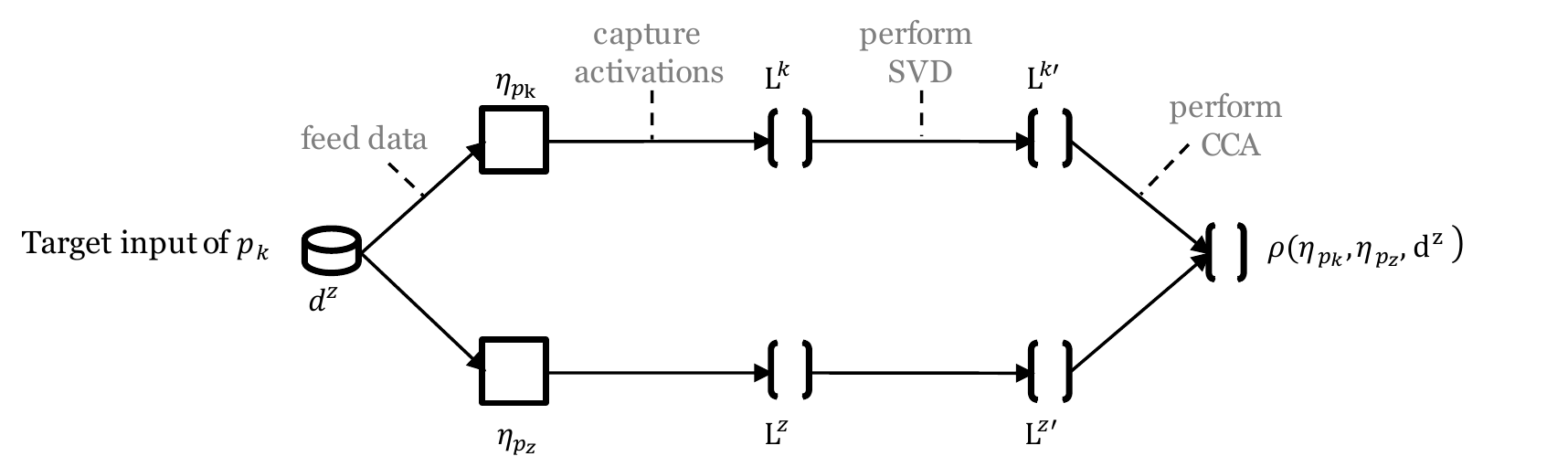}
    \caption{Procedure of comparing of comparing a potential source neural network $\eta_{p_z}$ to a target net $\eta_{p_k}$.}
    \label{fig:6-comparing}
\end{figure}

In this work, we use SVCCA to determine the neural net similarity $\rho$ of two networks $\eta_{p_k},\eta_{p_z}$ of two different branches. In Figure \ref{fig:6-comparing}, we present an overview of the application of SVCCA on a potential source net to identify its transferability to a target distribution. 

Two neural nets $\eta_{p_k},\eta_{p_z}$ that are to be compared are fed with data $d^z$. In this study we supply a data sample $d^z$ which represents the sales of 2017 from the target distribution to the potential source net and capture the activation vectors for every layer.

The neurons' response is calculated as a representation over a finite set of inputs. The resulting activation vectors $L^k$ for each layer of neurons are then processed by applying SVD. Similar to the eigenvalues, these characterize the properties of the matrix. This results in singular vectors $L^{k'}=(\{x'_1,...,x'_{m'_1}\})$ with associated singular values for X and similarly for Y. Of these singular vectors we keep the top $(m'_1)$, as 99\% of variation of X is explained by the top $(m'_1)$ vectors. This helps to remove directions with respect to neurons that are constantly zero or exhibit noise with small magnitudes \citep{Raghu2017SVCCA}.

Subsequently, CCA is applied to the sets of top singular vectors $(m'_1)$. The CCA is a well-established method for understanding the similarity of two different sets of randomly distributed variables. Given the two sets of vectors $(\{ x'_1,...,x'_{m'_1}\},\{ y'_1,...,y'_{m'_2}\})$, we wish to find linear transformations $(W_X,W_Y)$ that maximally correlate with the sub-spaces. This can be reduced to an eigenvalue problem. Solving this problem results in linearly transformed sub-spaces with directions $(\tilde{x}_i,\tilde{y}_i)$ that are maximally correlated with one another. As a result, we ultimately obtain $\rho = (\eta_{p_k},\eta_{p_z},d^z)$  as the transferability of a source neural net $\eta_{p_k}$  towards a target data set $\eta_{p_z}$.

\section{Results and Discussion}

We present the results of this study along two steps. First, we describe the result of the initial net training and the performed transfers---thus addressing H1. Second, we describe the output of the analysis on the association between data, data projection, neural net, and their impact on transferability---thus addressing H2-H4. 

\subsection{Base and Transfer Results (Hypothesis 1)}

To measure the performance of the developed forecasting models, two metrics are used: RMSE and MAPE. The RMSE is used to calculate the differences between values predicted by a model and the actual values observed and, in this work, is a basis during model optimization. It has proven to be a meaningful performance indicator for regression tasks \citep{Spuler2015}. RMSE is calculated as follows:
\begin{equation}
RMSE = \sqrt{\frac{\sum^T_{t=1} (\hat{y}_t-y_t)^2}{T}} 
\end{equation}
where $\hat{y}_t$ is the predicted value, $y_t$ is the actual value observed and T is the number of different predictions performed. RMSE as a scale-dependent measurement is not suitable for comparing forecasting errors across different data sets \citep{Hyndman2006AnotherAccuracy}. Thus, to evaluate and compare the performance of different models on different data sets, we additionally calculate the MAPE. The MAPE delivers a very intuitive interpretation in terms of relative error and therefore MAPE is broadly used in practice \citep{DeMyttenaere2016MeanModels} and is calculated as follows:
\begin{equation}
MAPE = \frac{100\%}{n} \sum^n_{t=1} |\frac{A_t-F_t}{A_t}|
\end{equation}
where $F_t$ is the forecast value and $A_t$ is the actual observation for the number of forecasts $n$. In the following, to ensure comparability, we only report the MAPE for all models.

\begin{figure}
    \centering
    \includegraphics[width=12cm]{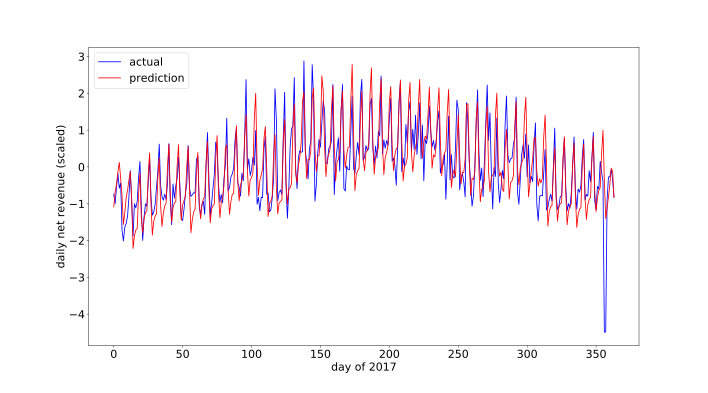}
    \includegraphics[width=12cm]{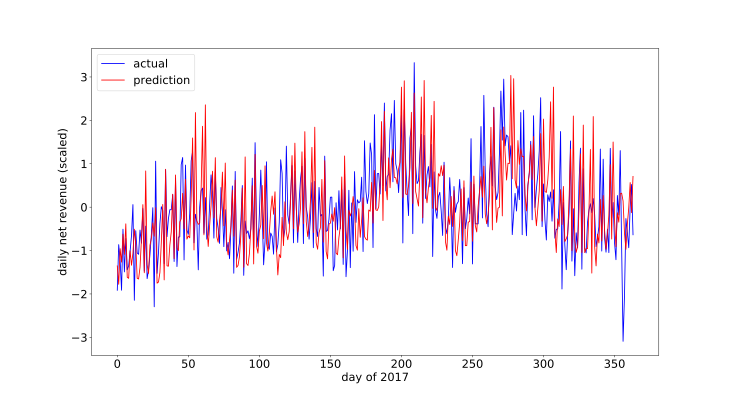}
    \caption{Scaled daily net revenue, actual and predicted; Above: branch 1, below: branch 4.}
    \label{fig:7-netrev}
\end{figure}

We train base models for every branch based on all available data including 2016. Then, we test the models on the full year of 2017 and calculate the MAPE and RMSE. In Figure \ref{fig:7-netrev}, we depict the scaled daily net revenue (exemplarily) for branch 1 and branch 4. Both base models are seemingly good in predicting the actual value. However, it is also noticeable that between those two data distributions---and, thus, models---there are significant differences in sales patterns.

As shown in Table \ref{tab:possible_transfers}, the number of potential transfers and therefore the number of possible models that are evaluated grows exponentially. However, to give an overview of the transfer impact, we present the results for the first degree of transfer in Table \ref{table:1stdegreeMAPE}.

\begin{table}
\centering
\resizebox{\textwidth}{!}{%
\begin{tabular}{|c|c|c|c|c|c|c|}
\hline
\diagbox{\textbf{Source $p_i$}}{\textbf{Target $p_i$}} & \multicolumn{1}{l|}{\textbf{base}} & \multicolumn{1}{l|}{\textbf{Branch 1 ($p_1$)}} & \multicolumn{1}{l|}{\textbf{Branch 2 ($p_2$)}} & \multicolumn{1}{l|}{\textbf{Branch 3 ($p_3$)}} & \multicolumn{1}{l|}{\textbf{Branch 4 ($p_4$)}} & \multicolumn{1}{l|}{\textbf{Branch 5 ($p_5$)}} \\ \hline
\begin{tabular}[c]{@{}l@{}}\textbf{-}\end{tabular} & \begin{tabular}[c]{@{}c@{}}9.59\end{tabular} & \begin{tabular}[c]{@{}c@{}}13.31\end{tabular} & \begin{tabular}[c]{@{}c@{}}13.94\end{tabular} & \begin{tabular}[c]{@{}c@{}}11.88\end{tabular} & \begin{tabular}[c]{@{}c@{}}23.00\end{tabular} & \begin{tabular}[c]{@{}c@{}}13.28\end{tabular} \\ 
\hline
\begin{tabular}[c]{@{}l@{}}\textbf{$\eta_{P_1}$}\end{tabular} & \begin{tabular}[c]{@{}c@{}}-\end{tabular} & \begin{tabular}[c]{@{}c@{}}13.13\\ (+1.34\%)\end{tabular} & \begin{tabular}[c]{@{}c@{}}15.23\\ (-9.30\%)\end{tabular} & \begin{tabular}[c]{@{}c@{}}11.14\\ (+6.28\%)\end{tabular} & \begin{tabular}[c]{@{}c@{}}25.51\\ (-10.91\%)\end{tabular} & \begin{tabular}[c]{@{}c@{}}13.42\\ (-1.23\%)\end{tabular} \\ \hline
\begin{tabular}[c]{@{}l@{}}\textbf{$\eta_{P_2}$}\end{tabular} & \begin{tabular}[c]{@{}c@{}}9.86\\ (-2.80\%)\end{tabular} & \begin{tabular}[c]{@{}c@{}}-\end{tabular} & \begin{tabular}[c]{@{}c@{}}13.85\\ (+0.67\%)\end{tabular} & \begin{tabular}[c]{@{}c@{}}10.64\\ (+10.44\%)\end{tabular} & \begin{tabular}[c]{@{}c@{}}24.78\\ (-7.74\%)\end{tabular} & \begin{tabular}[c]{@{}c@{}}13.96\\ (-5.29\%)\end{tabular} \\ \hline
\begin{tabular}[c]{@{}l@{}}\textbf{$\eta_{P_3}$}\end{tabular} & \begin{tabular}[c]{@{}c@{}}9.49\\ (+1.07\%)\end{tabular} & \begin{tabular}[c]{@{}c@{}}12.52\\ (+5.91\%)\end{tabular} & \begin{tabular}[c]{@{}c@{}}-\end{tabular} & \begin{tabular}[c]{@{}c@{}}11.21\\ (+5.67\%)\end{tabular} & \begin{tabular}[c]{@{}c@{}}24.71\\ (-7.46\%)\end{tabular} & \begin{tabular}[c]{@{}c@{}}13.07\\ (+1.38\%)\end{tabular} \\ \hline
\begin{tabular}[c]{@{}l@{}}\textbf{$\eta_{P_4}$}\end{tabular} & \begin{tabular}[c]{@{}c@{}}9.31\\ (+2.91\%)\end{tabular} & \begin{tabular}[c]{@{}c@{}}13.36\\ (-0.43\%)\end{tabular} & \begin{tabular}[c]{@{}c@{}}16.11\\ (-15.60\%)\end{tabular} & \begin{tabular}[c]{@{}c@{}}-\end{tabular} & \begin{tabular}[c]{@{}c@{}}25.59\\ (-11.29\%)\end{tabular} & \begin{tabular}[c]{@{}c@{}}12.82\\ (+3.31\%)\end{tabular} \\ \hline
\begin{tabular}[c]{@{}l@{}}\textbf{$\eta_{P_5}$}\end{tabular} & \begin{tabular}[c]{@{}c@{}}9.23\\ (+3.74\%)\end{tabular} & \begin{tabular}[c]{@{}c@{}}13.72\\ (-3.11\%)\end{tabular} & \begin{tabular}[c]{@{}c@{}}15.06\\ (-8.03\%)\end{tabular} & \begin{tabular}[c]{@{}c@{}}11.22\\ (+5.55\%)\end{tabular} & \begin{tabular}[c]{@{}c@{}}-\end{tabular} & \begin{tabular}[c]{@{}c@{}}13.11\\ (+1.10\%)\end{tabular} \\ \hline
\begin{tabular}[c]{@{}l@{}}$\eta_{P_6}$\end{tabular} & \begin{tabular}[c]{@{}c@{}}9.18\\ (+4.30\%)\end{tabular} & \begin{tabular}[c]{@{}c@{}}12.89\\ (+3.12\%)\end{tabular} & \begin{tabular}[c]{@{}c@{}}15.01\\ (-7.68\%)\end{tabular} & \begin{tabular}[c]{@{}c@{}}10.97\\ (+7.70\%)\end{tabular} & \begin{tabular}[c]{@{}c@{}}24.82\\ (-7.92\%)\end{tabular} & \begin{tabular}[c]{@{}c@{}}-\end{tabular} \\ \hline
\begin{tabular}[c]{@{}l@{}}\textbf{Best} \\ \textbf{Transfer}\end{tabular} & \begin{tabular}[c]{@{}c@{}}9.18\\ (+4.30\%)\end{tabular} & \begin{tabular}[c]{@{}c@{}}12.52\\ (+5.91\%)\end{tabular} & \begin{tabular}[c]{@{}c@{}}13.85\\ (+0.67\%)\end{tabular} & \begin{tabular}[c]{@{}c@{}}10.64\\ (+10.44\%)\end{tabular} & \begin{tabular}[c]{@{}c@{}}24.71\\ (-7.46\%)\end{tabular} & \begin{tabular}[c]{@{}c@{}}12.82\\ (+3.31\%)\end{tabular} \\ \hline
\end{tabular}}
\caption{MAPE M (the lower the better) results for the first degree of transfer of all branches. Additionally, the performance increase in comparison to a model that is trained solely on the target’s data (in brackets).}
\label{table:1stdegreeMAPE}
\end{table}

Not all transfers have a positive impact on the performance to a target distribution (see Table \ref{table:1stdegreeMAPE}). This indicates that transferability varies, depending on the association between the source and target distribution. Additionally, in practice it might not be feasible to test all possible transfer model candidates on a target set, as a transfer and re-training of a model is bound to a computational cost. Simply testing all possible combinations via a brute-force approach would therefore not be efficient.

\begin{table}
\centering
\resizebox{\textwidth}{!}{%
\begin{tabular}{|c|c|c|c|c|c|c|}
\hline
\diagbox{\textbf{Target $p_i$}}{\textbf{Degr. of T.}} & \multicolumn{1}{l|}{\textbf{base}} & \multicolumn{1}{l|}{\textbf{1$^st$ degree}} & \multicolumn{1}{l|}{\textbf{2$^nd$ degree}} & \multicolumn{1}{l|}{\textbf{3$^rd$ degree}} & \multicolumn{1}{l|}{\textbf{4$^th$ degree}} & \multicolumn{1}{l|}{\textbf{5$^th$ degree}} \\ \hline
\begin{tabular}[c]{@{}l@{}}Branch 1\\ ($p_1$)\end{tabular} & \begin{tabular}[c]{@{}c@{}}9.59\\ ($p_1$)\end{tabular} & \begin{tabular}[c]{@{}c@{}}9.18\\ ($p_6$,$p_1$)\end{tabular} & \begin{tabular}[c]{@{}c@{}}9.08\\ ($p_5$,$p_4$,$p_1$)\end{tabular} & \begin{tabular}[c]{@{}c@{}}8.98\\ ($p_3$,$p_5$,$p_4$,$p_1$)\end{tabular} & \begin{tabular}[c]{@{}c@{}}8.96\\ ($p_2$,$p_4$,$p_6$,$p_5$,$p_1$)\end{tabular} & \begin{tabular}[c]{@{}c@{}}8.96\\ ($p_5$,$p_6$,$p_3$,$p_2$,$p_4$,$p_1$)\end{tabular} \\ \hline
\begin{tabular}[c]{@{}l@{}}Branch 2\\($p_2$)\end{tabular} & \begin{tabular}[c]{@{}c@{}}13.31\\ ($p_2$)\end{tabular} & \begin{tabular}[c]{@{}c@{}}12.52\\ ($p_3$,$p_2$)\end{tabular} & \begin{tabular}[c]{@{}c@{}}11.87\\ ($p_6$,$p_3$,$p_2$)\end{tabular} & \begin{tabular}[c]{@{}c@{}}11.73\\ ($p_3$,$p_5$,$p_1$,$p_2$)\end{tabular} & \begin{tabular}[c]{@{}c@{}}11.65\\ ($p_3$,$p_4$,$p_6$,$p_1$,$p_2$)\end{tabular} & \begin{tabular}[c]{@{}c@{}}11.70\\ ($p_3$,$p_4$,$p_6$,$p_1$,$p_5$,$p_2$)\end{tabular} \\ \hline
\begin{tabular}[c]{@{}l@{}}Branch 3\\($p_3$)\end{tabular} & \begin{tabular}[c]{@{}c@{}}13.94\\ ($p_3$)\end{tabular} & \begin{tabular}[c]{@{}c@{}}13.84\\ ($p_2$,$p_3$)\end{tabular} & \begin{tabular}[c]{@{}c@{}}13.76\\ ($p_2$,$p_6$,$p_3$)\end{tabular} & \begin{tabular}[c]{@{}c@{}}13.38\\ ($p_2$,$p_6$,$p_1$,$p_3$)\end{tabular} & \begin{tabular}[c]{@{}c@{}}13.25\\ ($p_6$,$p_5$,$p_1$,$p_2$,$p_3$)\end{tabular} & \begin{tabular}[c]{@{}c@{}}13.01\\ ($p_2$,$p_6$,$p_4$,$p_5$,$p_1$,$p_3$)\end{tabular} \\ \hline
\begin{tabular}[c]{@{}l@{}}Branch 4\\ Co. B\end{tabular} & \begin{tabular}[c]{@{}c@{}}11.88\\ ($p_4$)\end{tabular} & \begin{tabular}[c]{@{}c@{}}10.64\\ ($p_2$,$p_4$)\end{tabular} & \begin{tabular}[c]{@{}c@{}}10.33\\ ($p_2$,$p_6$,$p_4$)\end{tabular} & \begin{tabular}[c]{@{}c@{}}10.18\\ ($p_3$,$p_1$,$p_2$,$p_4$)\end{tabular} & \begin{tabular}[c]{@{}c@{}}10.22\\ ($p_5$,$p_2$,$p_6$,$p_3$,$p_4$)\end{tabular} & \begin{tabular}[c]{@{}c@{}}10.03\\ ($p_2$,$p_6$,$p_5$,$p_3$,$p_1$,$p_4$)\end{tabular} \\ \hline
\begin{tabular}[c]{@{}l@{}}Branch 5\\($p_5$)\end{tabular} & \begin{tabular}[c]{@{}c@{}}23.00\\ ($p_5$)\end{tabular} & \begin{tabular}[c]{@{}c@{}}24.71\\ ($p_3$,$p_5$)\end{tabular} & \begin{tabular}[c]{@{}c@{}}23.19\\ ($p_3$,$p_6$,$p_5$)\end{tabular} & \begin{tabular}[c]{@{}c@{}}22.42\\ ($p_6$,$p_1$,$p_4$,$p_5$)\end{tabular} & \begin{tabular}[c]{@{}c@{}}22.16\\ ($p_1$,$p_6$,$p_4$,$p_3$,$p_5$)\end{tabular} & \begin{tabular}[c]{@{}c@{}}21.98\\ ($p_6$,$p_2$,$p_1$,$p_3$,$p_4$,$p_5$)\end{tabular} \\ \hline
\begin{tabular}[c]{@{}l@{}}Branch 6\\($p_6$)\end{tabular} & \begin{tabular}[c]{@{}c@{}}13.26\\ ($p_6$)\end{tabular} & \begin{tabular}[c]{@{}c@{}}12.82\\ ($p_4$,$p_6$)\end{tabular} & \begin{tabular}[c]{@{}c@{}}12.95\\ ($p_2$,$p_5$,$p_6$)\end{tabular} & \begin{tabular}[c]{@{}c@{}}12.42\\ ($p_3$,$p_1$,$p_5$,$p_6$)\end{tabular} & \begin{tabular}[c]{@{}c@{}}12.49\\ ($p_2$,$p_5$,$p_1$,$p_4$,$p_6$)\end{tabular} & \begin{tabular}[c]{@{}c@{}}12.21\\ ($p_2$,$p_3$,$p_1$,$p_5$,$p_4$,$p_6$)\end{tabular} \\ \hline
\end{tabular}}
\caption{MAPE M (the lower the better) of best model along degrees of transfer for each distribution $p_i$ with the corresponding transfer path in brackets.}
\label{table:nthdegreeMAPE}
\end{table}

In Table \ref{table:nthdegreeMAPE}, the best results for each branch according the degree of transfer are presented. Note that we select the best performing model for every transfer step and every branch. For almost all branches, except branch 5, an increase in prediction performance can be observed with an increasing degree of transfer. In case of branch 5, we observe an increase of performance starting after the third transfer. It is noticeable that the increase steadily grows for every transfer step, albeit in some cases marginally. 

With an increasing degree of transfer, we can observe that in some cases the same distributions are used to re-train models. If, for instance, we investigate target branch 1, we can observe that branch 4 and 5 seem to be good previous distributions to train a model on. However, as we always re-train the complete net, an information loss is likely to arise after multiple transfer steps. H1 states that a model $\eta_{p_k,p_z}$ which is pre-trained on a distribution $p_k$  and transferred to a distribution $p_z$ outperforms a model $\eta_{p_z} \iff \Delta M (\eta_{p_k},\eta_{p_k,p_z}) \geq 0$. Thus, a two-sided one-sample t-test for the mean of all transferabilities $\Delta M$ (N=1950) is conducted to test if the average transferability significantly deviates from zero. With a mean of 0.00894, a standard deviation of .06728 and a p-value <.0001, the test confirms that the average transferability is positive. Thus, H1 is supported. Although the mean of $\Delta M$ is only slightly above zero, Table \ref{table:nthdegreeMAPE} illustrates that there is a steady increase of performances with every further transfer step. However, in that scenario, the best performing models are cherry-picked. In reality, it would not be desirable to test all 1950 transferred models, e.g., due to computational cost. Thus, it is desirable to know in advance which models will perform best. This leads us to the study of association on transferability.

\subsection{Associations on Transferability (Hypotheses 2-4)}

Returning to our previously defined research gap, we aim to find an indicator of transferability between two data distributions without comparing them directly. By establishing and testing H1, we first show the utility of a transfer in our use case. Now, we empirically study the correlation between three influence factors on transferability: the data divergence (H2), the projected data divergence (H3) and the SVCCA (H4). 
For every hypothesis, we calculate Spearman’s rank correlation coefficient between the transferability and the corresponding indicator. The coefficient describes both the strength and the direction of the relationship. The Spearman correlation evaluates the monotonic relationship between the two continuous variables: transferability and the corresponding indicator. 
The results are presented in Table \ref{tab:correlations-transferability}. We split H3 into three sub-hypotheses corresponding to the differing data projection functions we examine: H3.1 corresponds to the T-SNE, H3.2 to the PCA and H3.3 to the MDS. For every hypothesis, we examine N=1950 transferred models.

\begin{table}
\centering
\begin{tabular}{rrrrrrrr}
  \hline
\textbf{H} & \textbf{Indicator for transferability between $\eta_{p_k}$ and $\eta_{p_z}$} & \textbf{$r_s$} \\ 
  \hline
H2 & Data divergence $D[p_k ||p_z]$ & -.4294*** \\
H3.1 & Projected data (T-SNE) divergence $D[f_{TSNE} (x^k)||f_{TSNE} (x^z)]$ & .0668** \\
H3.2 & Projected data (PCA) divergence $D[f_{PCA} (x^k)||f_{PCA} (x^z)]$ & -.2397*** \\
H3.3 & Projected data (MDS) divergence $D[f_{MDS} (x^k)||f_{MDS} (x^z)]$ & -.3101*** \\
H4 & Neural net similarity (SVCCA) $\rho (\eta_{p_z},\eta_{p_k,p_z},d^z)$ & -.2245*** \\
  \hline
  & \tiny{"*" means $p< .05$, "**" means $p< .01$, "***" and means $p< .001$.} & \\
\end{tabular}
\caption{Spearman correlation of all tested indicators for transferability.} 
\label{tab:correlations-transferability}
\end{table}

Although we do not intend to find indications on raw data as it might not be feasible in business networks due to data confidentiality reasons, we formulate H2 to investigate whether or not there is an association without any transformation of data. H2 states that the divergence of two distributions $p_k$ and $p_z$, described as  $D[p_k ||p_z]$, correlates with the transferability $\Delta M (\eta_{p_z} ,\eta_{p_k,p_z})$. Results of the study indicate that there is a significant negative association between the data divergence $D[p_k ||p_z]$ and the transferability $\Delta M (\eta_{p_z},\eta_{p_k,p_z})$ ($r_s$=-.4294, p<.0001).

By projecting data and thus masking confidential information, we state and test different techniques for transferability indicators through H3. Thus, H3 describes that the divergence of the projection of two distributions $f(x^k)$ and $f(x^z)$ described as  $D[f(x^k)||f(x^z)]$ correlates with the transferability $\Delta M (\eta_{p_z},\eta_{p_k,p_z})$. The sub-hypotheses H3.1-3.3 describe different projection functions, respectively. For H3.1, results indicate that there is a positive association between the projected data divergence $D_{TSNE} [f(x^k)||f(x^z)]$ based on the T-SNE projection and the transferability $\Delta M (\eta_{p_z},\eta_{p_k,p_z})$ ($r_s$=-.0668, p<.05). However, the Spearman’s rho is rather low which indicates a weak correlation between the two variables. In the case of H3.2, however, the results paint a clearer picture: a negative correlation between the projected data divergence $D_{PCA} [f(x^k)||f(x^z)]$ and the transferability $\Delta M (\eta_{p_z},\eta_{p_k,p_z})$ is present ($r_s$=-.3101, p<.0001). A similar situation can be observed by considering the results of H3.3, where we find an even higher negative correlation between the projected data divergence $D_{MDS} [f(x^k)||f(x^z)]$ based on MDS to the transferability $\Delta M (\eta_{p_z},\eta_{p_k,p_z})$ ($r_s$=-.3101, p<.0001). Based on the results for H3.1-3.3, we can derive that the PCA and the MDS are better aligned with the identified correlation between data divergence and transferability (H1), as the direction of their correlations towards the transferability is the same. Furthermore, in case of the T-SNE, we only see a weak positive monotonous association. 

Through the comparison, although not exposing raw, but projected data, a possible breach of confidential information is not unlikely, as certain characteristics of the original data distribution are still extractable from the projection. Thus, we state and test H4 to find indications for transferability by the result of the SVCCA, a measure for neural net similarity. In case of H4, we state that the output of a Singular Value Canonical Correlation Analysis $\rho (\eta_{p_z},\eta_{p_k,p_z},d^z)$ correlates with the transferability $\Delta M (\eta_{p_z},\eta_{p_k,p_z},d^z)$. Our tests show a similar result as for H2, H3.2 and H3.3. We find a significant negative association between the neural net similarity $\rho (\eta_{p_z},\eta_{p_k,p_z},d^z)$ and the transferability $\Delta M (\eta_{p_z}  ,\eta_{p_k,p_z})$ ($r_s$=-.2245, p<.0001).

In summary, we can reject the null hypothesis for H2-H4. However, we observe differences in the results for each tested association. There seems to be a clear negative correlation between the projected data divergence based on PCA and MDS and the transferability as compared to T-SNE. Here, we observe a positive correlation with a Spearman's rho value below .07 whereas PCA and MDS exhibit larger, yet negative Spearman's rho values. Hence, we observe the same direction of correlation between the net similarity and the transferability, which indicates stable results.

\section{Discussion}

A multitude of insightful results can be derived from the conducted empiric research. First and foremost, what sparks our interest the most is the observed dominant, negative correlation effect between the transferability and the data and data projection divergence and neural net similarity. Based on previous research, one would expect a positive correlation to be present \citep{Xiao2012a}. However, in the regarded case, we assume that a neural network benefits from divergent or different observations which are not available in previous training data. 

Additionally, in our case we consider sales data collected by different restaurants. Although the data sets originate from two different chains which serve different types of food, the underlying sales patterns might be quite comparable. Results indicate that the underlying data distribution cannot yet be learned by looking at an isolated data population. Thus, we hypothesize that if a neural net receives a larger amount of diverging observations as inputs, its generalization and hence its performance improve. 

\begin{figure}
    \centering
    \includegraphics[width=9cm]{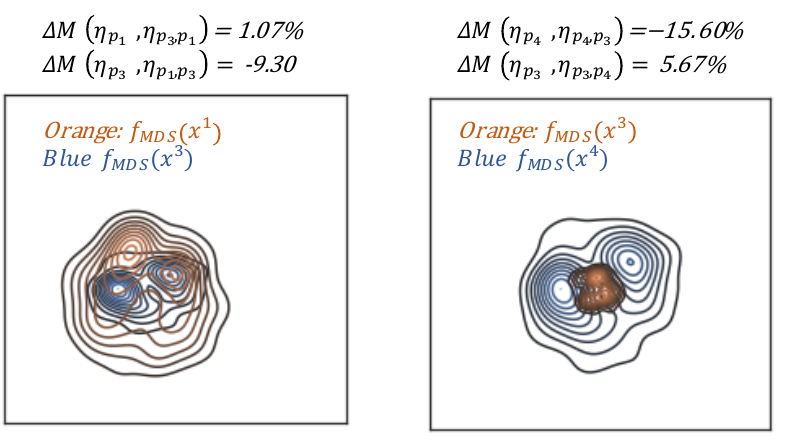}
    \caption{Overlay of bi-variate kernel density estimates of data projections (MDS) in the case of a) $f_{MDS} (x^1)$,$f_{MDS} (x^3)$ and b) $f_{MDS} (x^3)$, $f_{MDS} (x^4)$ and their respective bi-directional transferabilities.}
    \label{fig:8-overlay}
\end{figure}

Another striking finding can be observed by visually inspecting the projections of data populations and their respective transferability and divergences. Exemplarily, we consider projections derived through MDS and compare a first degree of transfer. In Figure \ref{fig:8-overlay}, we present two cases where the effect of projected data divergence and the transferability can be visually observed for particularly "successful" transfers and "unsuccessful" transfers. In the figure, we can detect a strong support for our hypothesis validation, as successful transfers occur when the data is extremely divergent and vice-versa, unsuccessful transfers occur when data is divergent. However, future work is necessary to further investigate this phenomenon. 

Furthermore, the correlations of the data and data projection divergence and their transferability show the same direction as the correlation between the neural net similarity and the transferability. This gives us reasons to believe that the neural net similarity, as applied in this work with SVCAA, represents similar abstracted information as the divergence of data and its projection. It also aligns with the work of  \citet{Raghu2017SVCCA}, who aim to find representations of features of a data set in a neuron’s response. However, this assumption requires further confirmation in future work based on additional empirical research established through other data sets.

\section{Conclusion and Outlook}

In this work, we utilize transfer machine learning on a unique sales data set. We do so to reveal two aspects of interest: first, the performance increase---labeled as transferability---of transferring models in general and second, the identification of indicators of a successful transfer prior to the transfer itself. 

Therefore, we contribute to the body of knowledge in manifold ways. First, we implement a multi-step system-wide transfer on the sales data of different restaurants and restaurant chains. We can successfully show the evaluation of utility by showing empirically that transfers can be beneficial. This is in line with Hypothesis 1, which states that a model that is pre-trained on one distribution and subsequently transferred to another distribution outperforms the model built solely on the original distribution. Secondly, the association of divergence of data distributions as well as the divergence of projections of data distributions and their transferability is analyzed. We are able to confirm Hypothesis 2 and Hypothesis 3 for different sub-distributions, indicating a strong negative correlation between data divergence and data projection divergence and their transferability. Thirdly, we analyze with Hypothesis 4 whether the output of a Singular Value Canonical Correlation Analysis is associated with the transferability. Although we analyze only trained nets---and not data distributions or their projections---we are able to find an association between the neural net similarity and the transferability. In summary, this means for the regarded data set that we are now able to determine transferability of models without regarding raw data---prior to the transfer. As a result, predictions about the transferability for new data sets in a business network can be made, without exposing data distributions. Additionally, its application could allow for more efficiency across the overall system, as the same problem does not need to be solved multiple times: a once trained model can be re-applied several times for similar problems at each restaurant.

Despite the novelty of the approach, limitations are obvious. At first, only one data set of multiple entities, only time series forecasting and only one net architecture is considered. In the theorizing process of general indicators for transferability, more examples are necessary. Additionally, for the time being, we only show an association between data, data projection and neural net similarity and the transferability. We do not investigate further and enhance the association to engineer a search algorithm for transferring models in an ecosystem. On the technical side, the currently implemented transfer mechanism exploits "forgetting", i.e., we do not dynamically adapt the frozen layers. Furthermore, the data and data projection association towards transferability neglects previous transfer steps of a model and is thus trivialized. Finally, while no raw data is shared, recent research shows the possibility to retrieve single instances, especially extreme points of a population \citep{fredrikson2015model}.

Future research needs to address especially the last aspect. If we aim to allow privacy-preserving transfer machine learning, we need to incorporate differential privacy mechanisms into model training \citep{abadi2016deep, Mironov2017}. Furthermore, the empirical study can be extended by incorporating previous training sets, as these could result in stronger correlations, e.g. due to averaging over populations. A further enhancement of the transfer mechanism could prove meaningful, for instance by including the freezing of certain layers, as well as adapting the learning rate or number of frozen layers with respect to the degree of transfer. Also, an in-depth investigation of the "forgetting" aspects of networks could be interesting, e.g., how many transfer steps are required for a network to "forget" information---and therefore limit the amount of transfers from the beginning. As mentioned previously, more and repeated empirical studies on other data sets, models, and net architectures are necessary to address the generalizability of the approach. Finally, an exploitation of the association between SVCCA and transferability would be preferable, specifically the development of a method or search algorithm that utilizes it as a direction of search. This would allow to choose the "path of transfer" in advance---and result in higher model performances with less model transfer permutations. A promising field of research lies ahead.

\bibliographystyle{unsrtnat}
\bibliography{references}  

\begin{thebibliography}{36}
\providecommand{\natexlab}[1]{#1}
\providecommand{\url}[1]{\texttt{#1}}
\expandafter\ifx\csname urlstyle\endcsname\relax
  \providecommand{\doi}[1]{doi: #1}\else
  \providecommand{\doi}{doi: \begingroup \urlstyle{rm}\Url}\fi

\bibitem[Sanders et~al.(2016)Sanders, Elangeswaran, and Wulfsberg]{Sanders2016}
Adam Sanders, Chola Elangeswaran, and Jens Wulfsberg.
\newblock {Industry 4.0 implies lean manufacturing: Research activities in
  industry 4.0 function as enablers for lean manufacturing}.
\newblock \emph{Journal of Industrial Engineering and Management}, 2016.
\newblock ISSN 20130953.
\newblock \doi{10.3926/jiem.1940}.

\bibitem[Hirt and K{\"{u}}hl(2018)]{Hirt2018}
Robin Hirt and Niklas K{\"{u}}hl.
\newblock Cognition in the era of smart service systems: Inter-organizational
  analytics through meta and transfer learning.
\newblock \emph{{Thirty Ninth International Conference on Information
  Systems}}, 2018.

\bibitem[Mizoguchi et~al.(1995)Mizoguchi, Vanwelkenhuysen, and
  Ikeda]{mizoguchi1995task}
Riichiro Mizoguchi, Johan Vanwelkenhuysen, and Mitsuru Ikeda.
\newblock {Task ontology for reuse of problem solving knowledge}.
\newblock \emph{Towards Very Large Knowledge Bases: Knowledge Building {\&}
  Knowledge Sharing}, 46\penalty0 (59):\penalty0 45, 1995.

\bibitem[Hicks(1939)]{hicks1939foundations}
J~R Hicks.
\newblock {The foundations of welfare economics}.
\newblock 1939.

\bibitem[Hopf et~al.(2017)Hopf, Sodenkamp, Riechel, and Staake]{Hopf2017a}
K.~Hopf, M.~Sodenkamp, S.~Riechel, and T.~Staake.
\newblock Predictive customer data analytics—the value of public statistical
  data and the geographic model transferability.
\newblock In \emph{{Proceedings of the International Conference on Information
  Systems}}, 2017.

\bibitem[Weiss et~al.(2016)Weiss, Khoshgoftaar, and Wang]{Weiss2016}
Karl Weiss, Taghi~M. Khoshgoftaar, and DingDing Wang.
\newblock {A survey of transfer learning}.
\newblock \emph{Journal of Big Data}, 3\penalty0 (1):\penalty0 9, dec 2016.
\newblock ISSN 2196-1115.
\newblock \doi{10.1186/s40537-016-0043-6}.
\newblock URL
  \url{http://journalofbigdata.springeropen.com/articles/10.1186/s40537-016-0043-6}.

\bibitem[Long et~al.(2013)Long, Wang, Ding, Sun, and Yu]{long2013transfer}
Mingsheng Long, Jianmin Wang, Guiguang Ding, Jiaguang Sun, and Philip~S Yu.
\newblock Transfer feature learning with joint distribution adaptation.
\newblock In \emph{{Proceedings of the IEEE International Conference on
  Computer Vision}}, pages 2200--2207, 2013.

\bibitem[Huh et~al.(2016)Huh, Agrawal, and Efros]{huh2016makes}
Minyoung Huh, Pulkit Agrawal, and Alexei~A Efros.
\newblock {What makes ImageNet good for transfer learning?}
\newblock \emph{arXiv preprint arXiv:1608.08614}, 2016.

\bibitem[Pan and Yang(2009)]{Pan2009ALearning}
Sinno~Jialin Pan and Qiang Yang.
\newblock A survey on transfer learning.
\newblock \emph{IEEE Transactions on Knowledge and Data engineering}, pages
  1345--1359, 2009.
\newblock \doi{10.1109/TKDE.2009.191}.
\newblock URL \url{http://socrates.acadiau.ca/courses/comp/dsilver/NIPS95}.

\bibitem[Bao et~al.(2019)Bao, Li, Huang, Zhang, Zamir, and
  Guibas]{Bao2019AnLearning}
Yajie Bao, Yang Li, Shao-Lun Huang, Lin Zhang, Amir~R. Zamir, and Leonidas~J.
  Guibas.
\newblock An information-theoretic metric of transferability for task transfer
  learning.
\newblock In \emph{{International Conference on Learning Representations (ICLR)
  2019}}, 2019.

\bibitem[Zhong et~al.(2010)Zhong, Fan, Yang, Verscheure, and
  Ren]{Zhong2010CrossLearning}
Erheng Zhong, Wei Fan, Qiang Yang, Olivier Verscheure, and Jiangtao Ren.
\newblock Cross validation framework to choose amongst models and datasets for
  transfer learning.
\newblock In Jose~Luis Balcazar and Francesco Bonchi, editors, \emph{{Machine
  Learning and Knowledge Discovery in Databases}}, pages 547--562, Berlin,
  Heidelberg, 2010. Springer Berlin Heidelberg.
\newblock ISBN 978-3-642-15939-8.

\bibitem[Kim(2014)]{Kim2014ConvolutionalClassification}
Yoon Kim.
\newblock Convolutional neural networks for sentence classification.
\newblock In \emph{{Conference on Empirical Methods in Natural Language
  Processing (EMNLP)}}, pages 1746--1751, Doha, Qatar, 2014. Association for
  Computational Linguistics.
\newblock URL \url{https://www.aclweb.org/anthology/D14-1181}.

\bibitem[Xue et~al.(2007)Xue, Liao, Carin, and
  Krishnapuram]{Xue2007Multi-TaskPriors}
Ya~Xue, Xuejun Liao, Lawrence Carin, and Balaji Krishnapuram.
\newblock {Multi-Task Learning for Classification with Dirichlet Process
  Priors}.
\newblock \emph{Journal of Machine Learning Research}, 8:\penalty0 35--63,
  2007.
\newblock URL \url{http://www.jmlr.org/papers/volume8/xue07a/xue07a.pdf}.

\bibitem[Yosinski et~al.(2014)Yosinski, Clune, Bengio, and
  Lipson]{Yosinski2014HowNetworks}
Jason Yosinski, Jeff Clune, Yoshua Bengio, and Hod Lipson.
\newblock {How transferable are features in deep neural networks?}
\newblock In \emph{{Advances in Neural Information Processing Systems 27}}.
  NIPS Foundation, 2014.
\newblock URL \url{https://arxiv.org/pdf/1411.1792.pdf}.

\bibitem[Jain and Learned-Miller(2011)]{Jain2011}
Vidit Jain and Erik Learned-Miller.
\newblock Online domain adaptation of a pre-trained cascade of classifiers.
\newblock In \emph{{Proceedings of the IEEE Computer Society Conference on
  Computer Vision and Pattern Recognition}}, 2011.
\newblock ISBN 9781457703942.
\newblock \doi{10.1109/CVPR.2011.5995317}.

\bibitem[Xiao et~al.(2012)Xiao, He, and Wang]{Xiao2012a}
Jin Xiao, Changzheng He, and Shouyang Wang.
\newblock Crude oil price forecasting: A transfer learning based analog
  complexing model.
\newblock In \emph{{2012 Fifth International Conference on Business
  Intelligence and Financial Engineering}}, pages 29--33. IEEE, aug 2012.
\newblock ISBN 978-1-4673-2092-4.
\newblock \doi{10.1109/BIFE.2012.14}.
\newblock URL \url{http://ieeexplore.ieee.org/document/6305073/}.

\bibitem[Bhattacharyya(1943)]{Bhattacharyya1943OnDistributions}
A~Bhattacharyya.
\newblock {On a measure of divergence between two statistical populations
  defined by their probability distributions}.
\newblock \emph{Bull. Calcutta Math. Soc.}, 35:\penalty0 99--109, 1943.

\bibitem[Eguchi(1985)]{Eguchi1985AFunctionals}
Shinto Eguchi.
\newblock {A differential geometric approach to statistical inference on the
  basis of contrast functionals}.
\newblock \emph{Hiroshima Mathematical Journal}, 15\penalty0 (2):\penalty0
  341--391, 1985.
\newblock ISSN 0018-2079.
\newblock \doi{10.32917/hmj/1206130775}.
\newblock URL \url{https://projecteuclid.org/euclid.hmj/1206130775}.

\bibitem[Kullback and Leibler(1951)]{Kullback1951OnSufficiency}
S~Kullback and R~A Leibler.
\newblock On information and sufficiency.
\newblock \emph{{The Annals of Mathematical Statistics}}, 22\penalty0
  (1):\penalty0 79--86, 1951.
\newblock ISSN 0003-4851.
\newblock \doi{10.1214/aoms/1177729694}.
\newblock URL \url{http://projecteuclid.org/euclid.aoms/1177729694}.

\bibitem[Raghu et~al.(2017)Raghu, Gilmer, Yosinski, and
  Sohl-Dickstein]{Raghu2017SVCCA}
Maithra Raghu, Justin Gilmer, Jason Yosinski, and Jascha Sohl-Dickstein.
\newblock {SVCCA:} singular vector canonical correlation analysis for deep
  learning dynamics and interpretability.
\newblock In \emph{{Advances in Neural Information Processing Systems 30}},
  pages 6076--6085. Curran Associates, Inc., 2017.
\newblock URL \url{https://arxiv.org/abs/1706.05806}.

\bibitem[Morcos et~al.(2018)Morcos, Raghu, and
  Bengio]{Morcos2018InsightsCorrelation}
Ari~S Morcos, Maithra Raghu, and Samy Bengio.
\newblock Insights on representational similarity in neural networks with
  canonical correlation.
\newblock In \emph{Advances in Neural Information Processing Systems 31}, pages
  5727--5736. Curran Associates, Inc., 2018.
\newblock URL \url{https://arxiv.org/abs/1806.05759}.

\bibitem[Borovykh et~al.(2017)Borovykh, Bohte, and
  Oosterlee]{Borovykh2017ConditionalNetworks}
Anastasia Borovykh, Sander Bohte, and Cornelis~W Oosterlee.
\newblock {Conditional time series forecasting with convolutional neural
  networks}.
\newblock In \emph{{Lecture Notes in Computer Science/Lecture Notes in
  Artificial Intelligence}}, pages 729--730, 2017.
\newblock URL \url{https://arxiv.org/abs/1703.04691}.

\bibitem[Kingma and Ba(2015)]{Kingma2015Adam:Optimization}
Diederik~P Kingma and Jimmy Ba.
\newblock Adam: A method for stochastic optimization.
\newblock In \emph{{International Conference On Learning Representations
  2015}}, pages 1--15. ICLR, 2015.
\newblock URL \url{http://arxiv.org/abs/1412.6980}.

\bibitem[Szab{\'{o}}(2014)]{Szabo2014InformationToolbox}
Zolt{\'{a}}n Szab{\'{o}}.
\newblock Information theoretical estimators toolbox.
\newblock \emph{{Journal of Machine Learning Research}}, 15:\penalty0 283--287,
  2014.

\bibitem[Sz{\'{e}}kely and Rizzo(2013)]{Szekely2013EnergyDistances}
G{\'{a}}bor~J Sz{\'{e}}kely and Maria~L Rizzo.
\newblock {Energy statistics: A class of statistics based on distances}.
\newblock \emph{Journal of Statistical Planning and Inference}, 143\penalty0
  (8):\penalty0 1249--1272, 2013.
\newblock \doi{10.1016/J.JSPI.2013.03.018}.
\newblock URL
  \url{https://www.sciencedirect.com/science/article/pii/S0378375813000633}.

\bibitem[Narayanan and Shmatikov(2008)]{Narayanan2008}
Arvind Narayanan and Vitaly Shmatikov.
\newblock Robust de-anonymization of large sparse datasets.
\newblock In \emph{{Proceedings - IEEE Symposium on Security and Privacy}},
  2008.
\newblock ISBN 9780769531687.
\newblock \doi{10.1109/SP.2008.33}.

\bibitem[Pezzotti et~al.(2017)Pezzotti, Lelieveldt, Maaten, H{\"{o}}llt,
  Eisemann, and Vilanova]{Pezzotti2017ApproximatedAnalytics}
N~Pezzotti, B~P~F Lelieveldt, L~v.~d. Maaten, T~H{\"{o}}llt, E~Eisemann, and
  A~Vilanova.
\newblock Approximated and user steerable tsne for progressive visual
  analytics.
\newblock \emph{{IEEE Transactions on Visualization and Computer Graphics}},
  23\penalty0 (7):\penalty0 1739--1752, 2017.
\newblock ISSN 1077-2626.
\newblock \doi{10.1109/TVCG.2016.2570755}.

\bibitem[Borg and Groenen(2003)]{Borg2003ModernApplications}
I~Borg and P~Groenen.
\newblock Modern multidimensional scaling: Theory and applications.
\newblock \emph{Journal of Educational Measurement}, 40\penalty0 (3):\penalty0
  277--280, 2003.
\newblock ISSN 0022-0655.
\newblock \doi{10.1111/j.1745-3984.2003.tb01108.x}.
\newblock URL \url{http://doi.wiley.com/10.1111/j.1745-3984.2003.tb01108.x}.

\bibitem[Halko et~al.(2011)Halko, Martinsson, and
  Tropp]{Halko2011FindingDecompositions}
N~Halko, P~G Martinsson, and J~A Tropp.
\newblock Finding structure with randomness: Probabilistic algorithms for
  constructing approximate matrix decompositions.
\newblock \emph{{SIAM Review}}, 53\penalty0 (2):\penalty0 217--288, 2011.
\newblock ISSN 0036-1445.
\newblock \doi{10.1137/090771806}.
\newblock URL \url{http://epubs.siam.org/doi/10.1137/090771806}.

\bibitem[Hardoon et~al.(2004)Hardoon, Szedmak, and
  Shawe-Taylor]{Hardoon2004CanonicalMethods}
David~R Hardoon, Sandor Szedmak, and John Shawe-Taylor.
\newblock Canonical correlation analysis: An overview with application to
  learning methods.
\newblock \emph{{Neural Computation}}, 16\penalty0 (12):\penalty0 2639--2664,
  2004.
\newblock ISSN 0899-7667.
\newblock \doi{10.1162/0899766042321814}.
\newblock URL
  \url{http://www.mitpressjournals.org/doi/10.1162/0899766042321814}.

\bibitem[Spuler et~al.(2015)Spuler, Sarasola-Sanz, Birbaumer, Rosenstiel, and
  Ramos-Murguialday]{Spuler2015}
Martin Spuler, Andrea Sarasola-Sanz, Niels Birbaumer, Wolfgang Rosenstiel, and
  Ander Ramos-Murguialday.
\newblock {Comparing metrics to evaluate performance of regression methods for
  decoding of neural signals}.
\newblock In \emph{{2015 37th Annual International Conference of the IEEE
  Engineering in Medicine and Biology Society (EMBC)}}, pages 1083--1086. IEEE,
  aug 2015.
\newblock ISBN 978-1-4244-9271-8.
\newblock \doi{10.1109/EMBC.2015.7318553}.
\newblock URL \url{http://ieeexplore.ieee.org/document/7318553/}.

\bibitem[Hyndman and Koehler(2006)]{Hyndman2006AnotherAccuracy}
Rob~J Hyndman and Anne~B Koehler.
\newblock {Another look at measures of forecast accuracy}.
\newblock \emph{International Journal of Forecasting}, pages 679--688, 2006.

\bibitem[{De Myttenaere} et~al.(2016){De Myttenaere}, Golden, {Le Grand}, and
  Rossi]{DeMyttenaere2016MeanModels}
Arnaud {De Myttenaere}, Boris Golden, B{\'{e}}n{\'{e}}dicte {Le Grand}, and
  Fabrice Rossi.
\newblock Mean absolute percentage error for regression models.
\newblock \emph{Neurocomputing}, 192:\penalty0 38--48, 2016.
\newblock \doi{10.1016/J.NEUCOM.2015.12.114}.
\newblock URL \url{www.datascience.net,}.

\bibitem[Fredrikson et~al.(2015)Fredrikson, Jha, and
  Ristenpart]{fredrikson2015model}
Matt Fredrikson, Somesh Jha, and Thomas Ristenpart.
\newblock Model inversion attacks that exploit confidence information and basic
  countermeasures.
\newblock In \emph{{Proceedings of the 22nd ACM SIGSAC Conference on Computer
  and Communications Security}}, pages 1322--1333. ACM, 2015.

\bibitem[Abadi et~al.(2016)Abadi, Chu, Goodfellow, McMahan, Mironov, Talwar,
  and Zhang]{abadi2016deep}
Martin Abadi, Andy Chu, Ian Goodfellow, H.~Brendan McMahan, Ilya Mironov, Kunal
  Talwar, and Li~Zhang.
\newblock Deep learning with differential privacy.
\newblock In \emph{Proceedings of the 2016 {ACM SIGSAC} {C}onference on
  {C}omputer and {C}ommunications {S}ecurity}, pages 308--318. ACM, 2016.

\bibitem[Mironov(2017)]{Mironov2017}
Ilya Mironov.
\newblock R{\'{e}}nyi differential privacy.
\newblock \emph{{Proceedings - IEEE Computer Security Foundations Symposium}},
  pages 263--275, 2017.
\newblock ISSN 19401434.
\newblock \doi{10.1109/CSF.2017.11}.

\end{thebibliography}


\end{document}